\documentclass[10pt,twocolumn,letterpaper]{article}

\usepackage{wacv}              %
\usepackage{pifont}%

\usepackage{graphicx}

\usepackage{multirow}
\usepackage{booktabs}
\usepackage{makecell}
\definecolor{wacvblue}{rgb}{0.21,0.49,0.74}
\definecolor{myorange}{RGB}{242, 133, 0}

\usepackage[pagebackref,breaklinks,colorlinks,allcolors=wacvblue]{hyperref}

\def\wacvPaperID{1736} %
\def\confName{WACV}
\def\confYear{2026}
\newcommand{\posNum}[1]{\small{\textit{\textcolor{blue}{$+$ #1}}}}

\title{SUGAR: A Sweeter Spot for Generative Unlearning of Many Identities}

\author{
Dung Thuy Nguyen$^{\dagger}$, 
Quang Nguyen$^{\ddagger}$, 
Preston K. Robinette$^{\dagger}$, 
Eli Jiang$^{\dagger}$, \\
Taylor T. Johnson$^{\dagger}$, 
Kevin Leach$^{\dagger}$\\
\vspace{2mm}
$^{\dagger}$Vanderbilt University \quad 
$^{\ddagger}$Rutgers University\\
\vspace{1mm}
{\tt\small \{dung.t.nguyen, preston.k.robinette, allison.z.jiang\}@vanderbilt.edu}\\
{\tt\small \{taylor.johnson, kevin.leach\}@vanderbilt.edu} \quad
{\tt\small quang.ng@rutgers.edu}
}
\newcommand{\methodName}[0]{\texttt{SUGAR}}
\newcommand{\update}[1]{\textcolor{black}{#1}}

\newcommand{\idMetric}[0]{\textsf{ID}}
\newcommand{\fidMetric}[0]{\textsf{FID}}
\newcommand{\smbf}[1]{\noindent\textbf{#1}}

\begin{document}
\maketitle

\begin{abstract}
Recent advances in 3D-aware generative models have enabled high-fidelity image synthesis of human identities. However, this progress raises urgent questions around user consent and the ability to remove specific individuals from a model's output space. We address this by introducing \methodName, a framework for \textit{scalable generative unlearning} that enables the removal of \textit{many identities} (simultaneously or sequentially) without retraining the entire model. Rather than projecting unwanted identities to unrealistic outputs or relying on static template faces, ~\methodName~learns a \textit{personalized surrogate latent} for each identity, diverting reconstructions to visually coherent alternatives while preserving the model’s quality and diversity. We further introduce a \textit{continual utility preservation objective} that guards against degradation as more identities are forgotten. \update{\methodName{} achieves state-of-the-art performance in removing up to 200 identities, while delivering up to a 700\% improvement in retention utility compared to existing baselines.} Our code is publicly available at \href{here}{https://github.com/judydnguyen/SUGAR-Generative-Unlearn}.
\end{abstract}
    
\section{Introduction}
\label{sec:intro}

Despite the undeniable benefits of generative AI across various domains such as image synthesis and content creation~\cite{liao2022text,zhang2022styleswin,zhao2022generative,lee2022exp}, growing concerns have emerged regarding their potential misuse and unintended consequences~\cite{zlateva2024conceptual,ferrara2024genai,zucca2025regulating}. In particular, generative models can inadvertently memorize and reproduce identifiable faces or proprietary images from their training data, posing significant risks to intellectual property rights and personal privacy~\cite{shokri2017membership,hu2022membership,dealcala2024comprehensive}.

\begin{figure}[t!]
    \centering
    \includegraphics[width=0.9\linewidth]{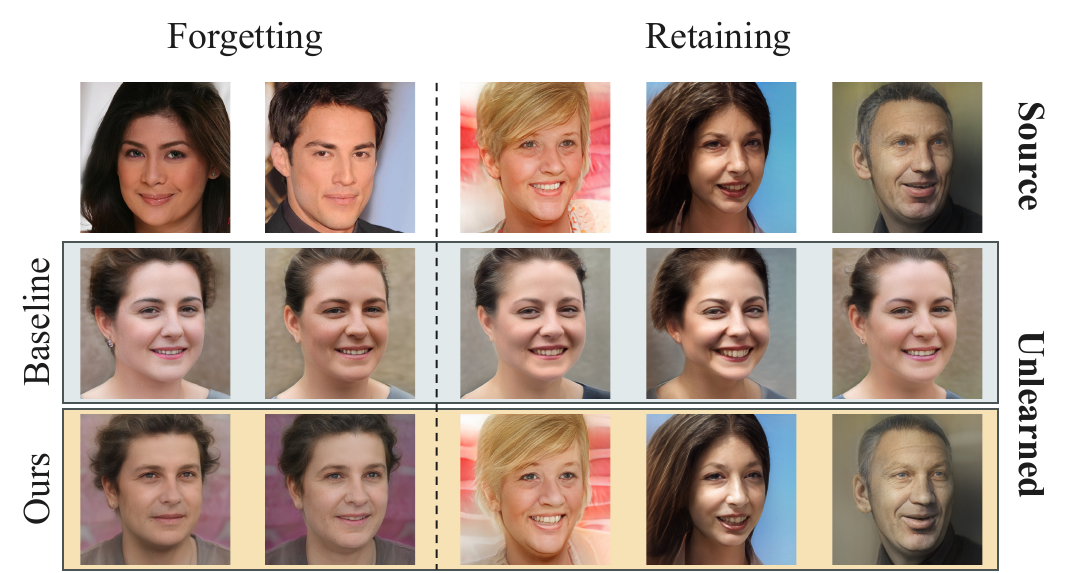}
    \caption{
    Results on forgetting multiple identities (left column). Our method effectively removes the specified identities from the generative model while maintaining the quality and distinctiveness of the retained identities. In contrast, the baseline (GUIDE) method causes noticeable degradation in the retained identities.}
    \label{fig:prelim-result}
    \vspace{-0.5cm}
\end{figure}

These issues have urged the \textit{right to be forgotten}~\cite{rosen2011right,zhang2024right}, 
a movement supporting individuals' ability to request the removal of personal data (e.g., images of one's face) from digital systems like AI models.
While possible to remove requested images from training sets, retraining models from scratch is computationally impractical, especially if there are multiple requests over time~\cite{li2025machine,yao2025large}.

To address the growing need for post-hoc data removal, recent work has explored \textit{generative unlearning}—the task of removing specific information from pretrained generative models without retraining from scratch~\cite{bourtoule2021machine,zhang2023review,heng2024selective,seo2024generative}. Approaches vary in how they suppress unwanted identities. Selective Amnesia~\cite{heng2024selective} targets diffusion and VAE-based models, using continual learning techniques to push erased concepts into noise-like outputs. GUIDE~\cite{seo2024generative}, the first to address \textit{Generative Identity Unlearning} (GUI) in GANs, instead manipulates the latent space by redirecting target embeddings toward a fixed mean identity representation. As illustrated in \autoref{fig:prelim-result}, both methods demonstrate initial effectiveness for single-identity removal but face key limitations: (a) they lack support for forgetting multiple identities at once or over time; (b) their suppression mechanisms—mapping to noise or an average face—may leak information about the forgotten identity; and (c) they often degrade unrelated generations, causing semantic drift in retained identities. For example, in GUIDE, retained faces in the Baseline row of \autoref{fig:prelim-result} begin to converge toward the median face, reducing output diversity and model utility.

Toward this end, we present~\methodName{}, which is capable of securely removing multiple identities while maintaining the usability of the model. 
Unlike previous approaches that introduce artifacts or distortions, \methodName{} ensures that forgotten identities are effectively erased without compromising the fidelity, diversity, or usability of the model's outputs. The contributions of this work, therefore, are the following:

\begin{enumerate}
    \item We introduce \methodName, which---to the best of our knowledge---is the first GUI method to address \textit{multiple-identity unlearning} and \textit{sequential unlearning} within the context of generative AI.
    \item We demonstrate the effectiveness of our approach over SOTA methods using Identity Similarity (\idMetric{}) and Fréchet Inception Distance (\fidMetric{}) quantitative metrics, \textbf{achieving up to a 700\% improvement over baselines in maintaining model utility}. Furthermore, our qualitative analysis and human judgment study reinforce these findings, showing that our method effectively unlearns multiple target identities while preserving the visual quality of retained identities. 
    \item We conduct extensive ablation studies to evaluate controllable unlearning, identity retention, utility preservation, and privacy enhancement, highlighting our method's advantages: (i) effectiveness in forgetting and retention; (ii) learnable and automatic determination of new identities for each forgotten identity; and (iii) privacy preservation throughout the unlearning process.
\end{enumerate}

\section{Related Works}
\label{sec:related}

\smbf{Mitigating Misuse of Generative Models. }
As generative models continue to advance in their ability to synthesize and manipulate highly realistic images, concerns about ethical misuse have intensified. The integration of powerful image editing techniques—such as latent space manipulation~\cite{shen2020interfacegan}, language-guided editing~\cite{patashnik2021styleclip}, and photorealistic face retouching~\cite{yoo2019photorealistic}—has made it increasingly accessible for users to modify images of real individuals or apply stylistic transformations~\cite{somepalli2023diffusion}. More alarmingly, recent studies have shown that these models can be exploited to extract memorized, copyrighted training data~\cite{carlini2023extracting}, often without the consent of the individuals or content owners involved. One of the most pressing risks is the unauthorized synthesis of a person’s likeness in misleading or harmful contexts, such as deepfakes or defamatory media~\cite{abbas2024unmasking,liu2024metacloak,romero2024generative}.
To counter this, Thanh et al.~\cite{van2023anti} developed an adversarial attack against UNet, effectively corrupting generated images.
Other works in the concept erasure domain focus on eliminating certain visual concepts.
For example, Gandikota et al.~\cite{gandikota2023erasing} were among the first to introduce removing specific concepts from diffusion models using negative guidance.
This is still an ongoing research direction with follow-up works such as~\cite{gandikota2023erasing,lu2024mace,petsiuk2024concept} to better improve the balancing between unlearning efficiency and maintaining model utility on remaining concepts.
Previous methods are identity-specific, requiring tailoring for each individual, which makes them non-scalable for protecting multiple identities and introduces substantial computational overhead. Moreover, these approaches primarily focus on associations between concepts, such as the ``Van Gogh style,'' and generated images, a focus more relevant to multi-modal learning, like text-to-image models. In contrast, our study centers on image-to-image models.

\smbf{Generative Unlearning. }
Machine unlearning is a technique that enables a machine learning model to ``forget'' or remove the influence of specific data points from its training data, without needing to be retrained from scratch~\cite{tiwary2023adapt,bourtoule2021machine,zhang2023review}. 
The usage of approximate unlearning is more preferable in the context of GenAI because these models are often trained on large corpora and datasets, making the retraining approach infeasible~\cite{xu2024machine,li2025single}. 
The interest in machine unlearning with generative models has grown recently, where the field was dominated by supervised unlearning before, i.e., classification tasks~\cite{gupta2021adaptive, tarun2023fast, golatkar2020eternal, golatkar2021mixed, chundawat2023zero, golatkar2020forgetting}. 
Seo et al.~\cite{seo2024generative} were the first to address GIU for GAN-based models in an approach called GUIDE. 
However, their method does not account for practical scenarios involving \textbf{(i) simultaneous unlearning requests}---an essential requirement in real-world applications where multiple individuals may request data removal at once---or \textbf{(ii) sequential unlearning}, where requests arrive at different times~\cite{zhang2024unlearncanvas}. 
Our empirical results also demonstrate that while GUIDE is effective for unlearning, it 
hampers the model's ability to maintain overall generation quality on retained identities.

Recent studies have also addressed generative unlearning by mapping forgotten identities to a Gaussian distribution $\mathcal{N}(0, \sigma)$ while preserving the original distribution for remaining data in an approach called Selective Amnesia (SA)~\cite{heng2024selective, feng2025controllable}. However, using Gaussian noise to replace forgotten identities is suboptimal, as it fails to accurately represent certain training data distributions, potentially corrupting other identities. Furthermore, SA, as well as GUIDE, \textbf{(iii) raise security concerns}, as the association between a forgotten identity and its output (noise or average face) can still be detected.
To this end, we introduce \methodName{} to address each of these concerns (\textbf{i}, \textbf{ii}, and \textbf{iii}).

\section{\methodName: Architecture and Components}
\label{sec:method}

\begin{figure*}[tbh]
    \centering
    \begin{minipage}{0.6\textwidth}
    \includegraphics[width=1.0\linewidth]{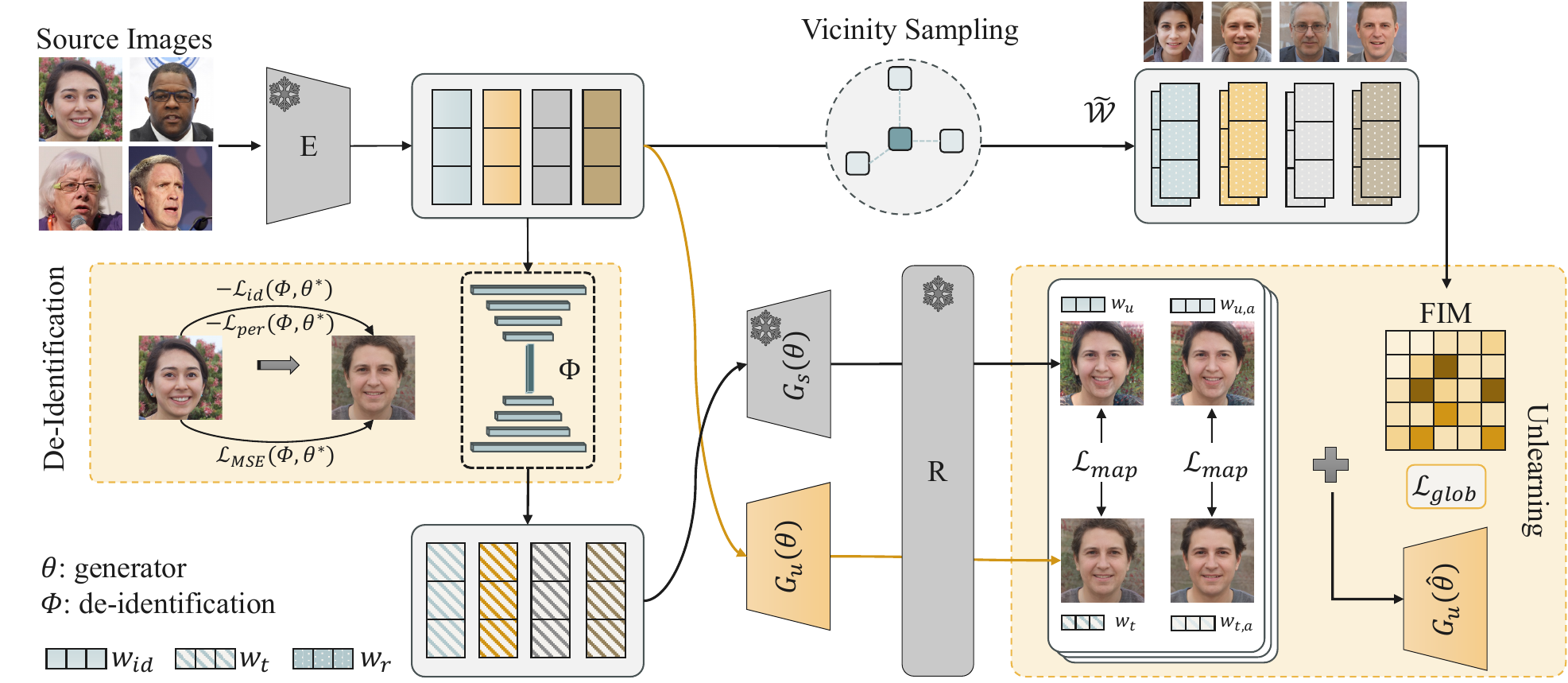}
    \caption{The overall pipeline of \methodName{} consists of two phases: (i) ID-specific De-Identification and (ii) Forgetting in a Continual Learning Fashion. In the first phase, a mapping function, $\Phi$, is trained to determine how a forgetting identity should be mapped using the forgetting set $\mathcal{D}_f$ and the pre-trained model $G_s$. In the second phase, the generator $\theta$ of the generative model is updated such that the forgotten identities are mapped to new, targeted identities generated by $\Phi$. 
    }
    \vspace{-0.3cm}
    \label{fig:overall-pipeline}
    \end{minipage}\hfill
     \begin{minipage}{0.35\textwidth}
     \includegraphics[width=1.0\textwidth]{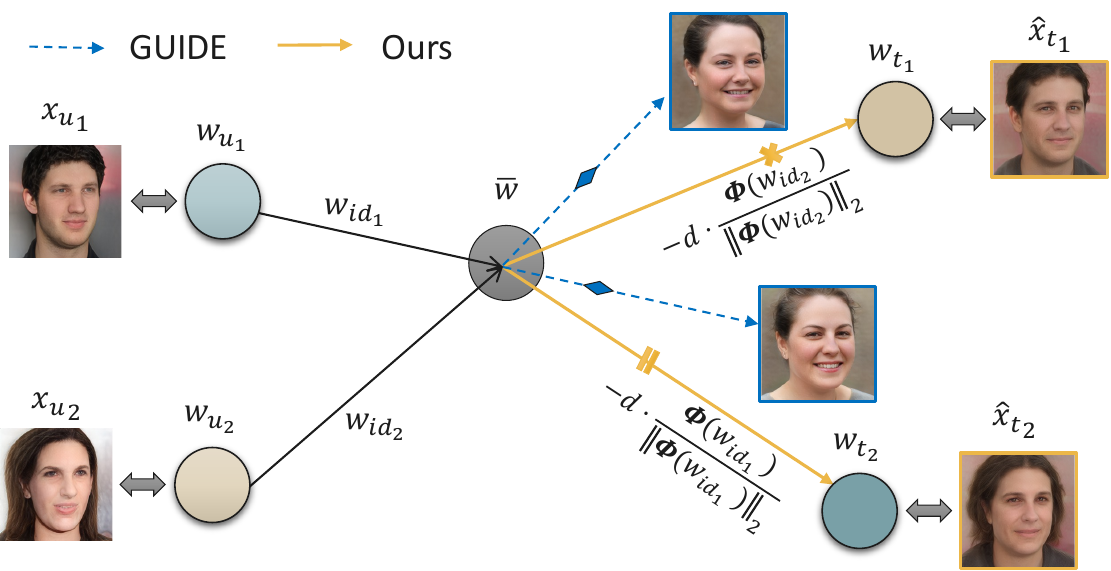}
    \caption{Our de-identification process, denoted as $\Phi$, defines the target latent code $w_t$ for unlearning by linearly subtracting the average identity vector $\Bar{w}$ from the counterfactual identity $\Phi(w_{id})$ produced by the model. This approach contrasts with GUIDE, which sets the new target vector by extrapolating from the source latent code along the direction toward $\Bar{w}$, maintaining a fixed distance along that direction.}
    \label{fig:deindent}
    \vspace{-0.5cm}
     \end{minipage}
\end{figure*}

\subsection{Problem Formulation}

\noindent\textbf{Problem formulation (multi-identity unlearning).}
We build on the GUI setup from Seo et al.~\cite{seo2024generative} and extend it to handle multiple identity unlearning. We consider a pretrained EG3D-style pipeline $(E_\psi, G_\theta, R)$ and aim to obtain an \emph{unlearned} generator $G_\theta^{u}$ that forgets a set of $K$ target identities while retaining performance on others.
Let $\mathcal{D} = \mathcal{D}_u \cup \mathcal{D}_r$ denote the full dataset, where $\mathcal{D}_u$ contains images of the $K$ identities to remove and $\mathcal{D}_r$ the remainder ($|\mathcal{D}_r|\!\gg\!|\mathcal{D}_u|$).
We operate in latent space: $W_u := \{\, w = E_\psi(x) \mid x \in \mathcal{D}_u \,\}$ and $W_r := \{\, w = E_\psi(x) \mid x \in \mathcal{D}_r \,\}$.
We keep the mapping network and renderer $R$ fixed and update only $G$.

For brevity, we define the feature operator $\mathcal{F}(w) := G(w)$, which returns the tri-plane feature, 
and the reconstruction operator $\mathcal{R}(w) := R(\mathcal{F}(w); c)$, which renders the final image. Accordingly, the objectives of multiple-identity unlearning can be formulated as follows.
\begin{itemize}[leftmargin=1em, itemsep=1pt]
\item \textit{Forget:} For every $w \in W_u$, the image $\hat{x}^{u}=\mathcal{R}_{G^u}(w)$ should not resemble the identity associated with $\hat{x}^{*}=\mathcal{R}_{G^*}(w)$.
\item \textit{Retain:} For every $w \in W_r$, $\hat{x}^{u}$ should remain close to $\hat{x}^{*}$ under standard perceptual/identity/photometric metrics.
\end{itemize}
\noindent\textbf{Remapping function.}
To disassociate image generation from an identity $\mathcal{I}_u$ with latent vector $w_u$, a straightforward and effective strategy—used in \textbf{GUIDE}~\cite{seo2024generative}—is to remap $w_u$ to a surrogate target identity $I^{avg}$ (e.g., the median face) with latent vector $w_{avg}$. With \textbf{GUI}, we adopt this idea and denote the GUIDE loss as the \emph{remapping objective}.
\begin{equation}
\label{eq:remap}
\begin{aligned}
& \mathcal{L}_{map}(w_u, w_{avg})
= \lambda_{mse}\,\mathcal{L}_{mse}\!\big(\mathcal{F}(w_u), \mathcal{F}(w_{avg})\big) \\
&\quad+ \lambda_{per}\,\mathcal{L}_{per}\!\big(\mathcal{R}(w_u), \mathcal{R}(w_{avg})\big)
+ \lambda_{id}\,\mathcal{L}_{id}\!\big(\mathcal{R}(w_u), \mathcal{R}(w_{avg})\big).
\end{aligned}
\end{equation}

This couples an MSE term in feature space with perceptual and identity terms in image space, aligning intermediate features and steering the generator toward the surrogate identity $I^{avg}$ rather than the original $\mathcal{I}_u$.

However, direct remapping can induce non-smooth behavior across multiple forgotten identities and trade off with retention.
To enable scalable removal, we introduce \textbf{ID-specific De-Identification}, which assigns a personalized surrogate target to each identity in $W_u$ and guides $G_\theta^{u}$ toward these substitutes while explicitly enforcing the retain constraint on $W_r$.

\subsection{ID-specific De-indentification}
Unlike existing unlearning approaches with predefined target concepts, we propose a \textbf{learnable} and \textbf{sample-specific} procedure to determine where each forgotten sample should be mapped. We call this \textit{ID-specific De-identification}.

Following Seo et al.~\cite{seo2024generative}, we define the identity latent vector \( w_{id} \) for an individual as the difference between their latent representation \( w_u \) and the average latent vector \( \bar{w} \), i.e., \( w_{id} = w_u - \bar{w} \). In our method, we train a mapping model \( \mathcal{T}_{\Theta} \) to produce a new identity vector \( w_{id}^t \) given any input latent vector \( w_u \in \mathbb{R}^{d \times h \times w} \), such that:
\begin{equation}
    w_{id}^t = \mathcal{T}_{\Theta}(w_{id}) \in \mathbb{R}^{d \times h \times w}
    \label{eqn:new-id}
\end{equation}

We then define the \textit{de-identification operator} \( \Theta(\cdot, w_{u}) \), which determines the target latent code \( w_t \) for the forgotten identity \( w_u \), by shifting the reference latent \( \bar{w} \) along a direction derived from \( w_{id} \). Specifically, the operator subtracts a normalized transformation of \( w_{id} \), scaled by a factor \( d \in \mathbb{R}^+ \), from \( \bar{w} \):
\begin{equation}
    \Theta(\cdot, w_{u}) := w_t = \bar{w} - d \cdot \frac{\mathcal{T}_{\Theta}(w_{id})}{\|\mathcal{T}_{\Theta}(w_{id})\|}
    \label{eqn:transformation}
\end{equation}
Here, \( d \) controls the displacement magnitude in latent space to ensure that the target code \( w_t \) removes identifiable traits of the original identity.

The intuition is that we learn a mapping $\mathcal{T}_\Theta$ that assigns each forgotten identity $w_u \in W_u$ a personalized \emph{counterfactual} $w_t$: a different identity that (i) preserves coarse facial attributes of $w_u$ while (ii) being recognizably distinct by facial recognition metrics such as ID or LPIPS~\cite{seo2024generative}. In latent space, $w_t$ serves as the surrogate for $w_u$; after unlearning, $G_\theta^u$ synthesizes faces consistent with $w_t$ rather than $w_u$.
To learn this identity replacement, we train a transformation function $\Theta$ that maps each $w_u \in W_u$ to a replacement latent code $w_t = \Theta(\cdot, w_u)$ that meets the above criteria. The transformation is supervised using the following objective:
\begin{multline}
\label{eqn:loss-de}
\mathcal{L}_{de}(\Theta)
= \frac{1}{|W_u|}\sum_{w_u \in W_u}\Big[
\lambda_{mse}\,\mathcal{L}_{mse}\!\big(\mathcal{F}(w_u),\,\mathcal{F}(\Theta(w_u))\big) \\
- \lambda_{per}\,\mathcal{L}_{per}(x_u, \hat{x}_u)
- \lambda_{id}\,\mathcal{L}_{id}(x_u, \hat{x}_u)
\Big]
\end{multline}
where the first term, $\mathcal{L}_{mse}$, encourages the replacement latent code to retain similar tri-plane features to $w_u$, ensuring that the substitute identity remains structurally plausible. The last two terms—$\mathcal{L}_{per}$ (perceptual loss) and $\mathcal{L}_{id}$ (identity loss)—enforce dissimilarity in image space between the reconstructed output $\hat{x}_u = \mathcal{R}(\Theta(w_u))$ and the original image $x_u$, promoting effective identity removal.

As shown in~\autoref{fig:deindent}, compared to the baseline GUIDE, our method introduces a more flexible mapping function, enabling the identification of a new target identity that better facilitates the subsequent unlearning process.  

\subsection{Forgetting in a Continual Learning Fashion}
In unlearning, there are always two objectives tied together, which are enforcing forgetting and retention.

\smbf{Enforcing Identity Forgetting.}
Building on the remapping objective of Seo et al.~\cite{seo2024generative} in~\autoref{eq:remap}, we replace the fixed median-face target with a \emph{learnable} de-identification target produced by our operator $\Theta(\cdot)$; that is, the average face $\mathcal{I}^{avg}$ is replaced by $\Theta(w)$.
As noted in the GUIDE baseline, successful unlearning must also suppress identity information in the local neighborhood of each forgotten latent. 
We therefore sample a perturbed neighbor $w_{u,a}$ for each forgotten code $w_u$:
\[
w_{u,a} \;=\; w_u \;+\; \alpha_n \,\frac{w_{r,a} - w_u}{\big\|w_{r,a} - w_u\big\|_2},
\qquad
\alpha_n \sim \mathcal{U}(0,\alpha_{\max}),
\]
where $w_{r,a}$ is a randomly chosen reference latent that supplies a direction; this produces marginal perturbations in the vicinity of $w_u$ to ensure robustness across local variations.
We apply the de-identification operator $\Theta$ to both the anchor code $w_u$ and its latent-space neighbor $w_{u,a}$, producing the learnable targets $\Theta(w_u)$ and $\Theta(w_{u,a})$. We then extend the remapping objective in~\autoref{eq:remap} to both targets, defining the overall forgetting loss:
\begin{equation}
\label{eqn:forget-loss}
\begin{aligned}
\mathcal{L}_{forget}(w_u;\theta)
&=
\mathcal{L}_{map}\!\big(w_u,\Theta(w_u)\big) \\
&\quad + \lambda_{nei}\,
\mathcal{L}_{map}\!\big(w_{u,a},\Theta(w_{u,a})\big).
\end{aligned}
\end{equation}
For brevity and further ablation study, we denote the second term as the neighbor-remapping loss,
$\mathcal{L}_{nei} := \lambda_{nei}\,
\mathcal{L}_{map}\!\big(w_{u,a},\Theta(w_{u,a})\big)$.

\smbf{Maintaining Model Utility.}  
During unlearning, we observe that samples whose latent representations are close to those in the forgetting set \( W_u \) are more likely to experience performance degradation, whereas samples farther away are typically unaffected
\footnote{See Appendix D.3 for further analysis.}. 
To preserve the model's utility, we propose a strategy to protect these nearby samples from unintended disruption.

\noindent\textit{Vicinity Sampling.}
Unlearning a set of identities can disrupt nearby latents belonging to other identities; in particular, identities whose codes lie close to the forgotten set are most prone to collateral degradation 
(cf. Appendix D.3).
To proactively guard against this effect, we construct a \emph{vicinity set} \( \Tilde{W} \) of perturbation probes around each forgotten latent and use them as the retention objective.
Formally, let $d^i \sim \mathcal{N}(0, I)$ be a random direction and $\hat{d}^i := d^i / \|d^i\|_2$ the corresponding unit vector. 
For a step radius $\alpha_r > 0$, we define the probe set:
\begin{equation}
\widetilde{W} \;:=\; \Big\{\, \widetilde{w}^{\,i} \;\Big|\; \widetilde{w}^{\,i} \;=\; w_u^i \;+\; \alpha_r\, \hat{d}^i \,\Big\}_{i=1}^{K}.
\label{eq:probe-set}
\end{equation}
Intuitively, $\widetilde{W}$ samples points on a small spherical shell centered at $w_u^i$. 
These probes are incorporated into the retention objective to prevent unlearning $W_f$ from inadvertently distorting neighboring latents and, in turn, degrading identities not intended to be forgotten.

While one might preserve nearby samples by adding explicit reconstruction or retain losses (e.g., enforcing the same reconstruction quality via~\autoref{eqn:forget-loss}), this introduces two drawbacks: 
(1) the computational cost grows linearly with the number of protected samples, becoming infeasible at scale; and 
(2) directly constraining generation can interfere with effective unlearning.

\noindent\textit{Elastic Weight Consolidation (EWC).}
To efficiently preserve behavior in regions most susceptible to collateral damage, we adopt Elastic Weight Consolidation (EWC)~\cite{kirkpatrick2017overcoming}. 
We treat the vicinity probe set $\widetilde{W}$ as a proxy for prior behavior to be retained and regularize parameters that are important for agreement with the source model around forgotten latents.

Let $\theta^*$ denote the pre-unlearning generator parameters. 
We define a reference (retain) loss that measures alignment with the source model on a probe $w \in \widetilde{W}$: $\ell_{\text{ref}}(w;\theta) := \mathcal{L}_{map}(\widetilde{w}^{\,i}, \widetilde{w}^{\,i}),\ \forall\, \widetilde{w}^{\,i} \in \widetilde{W}.$
We approximate the diagonal Fisher information over $\widetilde{W}$ at $\theta=\theta^*$ by the average squared gradient:
\begin{equation}
\label{eq:fisher-probe}
FIM_i \;\approx\; \frac{1}{|\widetilde{W}|}\sum_{w\in \widetilde{W}}
\left( \frac{\partial\, \ell_{\text{ref}}(w;\theta)}{\partial \theta_i}\Big|_{\theta=\theta^*} \right)^{\!2},
\end{equation}
computed with a mini-batch estimator in practice.
Given these importances, the EWC penalty constrains drift from $\theta^*$ proportionally to $F_i$:
\begin{equation}
\label{eq:ewc}
\mathcal{L}_{ewc} \;=\; \frac{1}{2}\sum_{i} F_i\,\big(\theta_i - \theta_i^*\big)^2.
\end{equation}
This regularizer encourages parameters most critical for preserving behavior on the \emph{vicinity} set to remain near their pre-unlearning values, thereby avoiding per-sample retain losses during unlearning and reducing collateral distortion around forgotten identities.

\noindent\textit{Final Objective.}  
Combining the forgetting loss over the identities in \( W_u\) with the EWC regularization, the final unlearning objective is:
\begin{equation}
\label{eqn:loss-full}
\mathcal{L}_{unlearn}
=
\mathbb{E}_{\,w_u \sim W_u}\!
\big[\,\mathcal{L}_{forget}(w_u;\theta)\,\big]
\;+\;
\lambda_{ewc}\,
\mathcal{L}_{ewc}(\theta;\theta^*).
\end{equation}

This formulation enables us to mitigate side effects on samples near the forgetting set while preserving overall model utility, without incurring the high cost of sample-specific loss terms.

Overall, \methodName{} effectively forgets identities in a pre-trained EG3D generator by (1) leveraging a new ID-specific DeIdentification process that maps a target identity to forget to a new identity in the latent space, and (2) combining forgetting and retaining loss function.  
This approach ensures identities can be effectively removed while ensuring model utility on other identities.    

\section{Experiments}

\label{sec:experiments}
\begin{figure*}[tbh]
    \centering
    \includegraphics[width=0.8\linewidth]{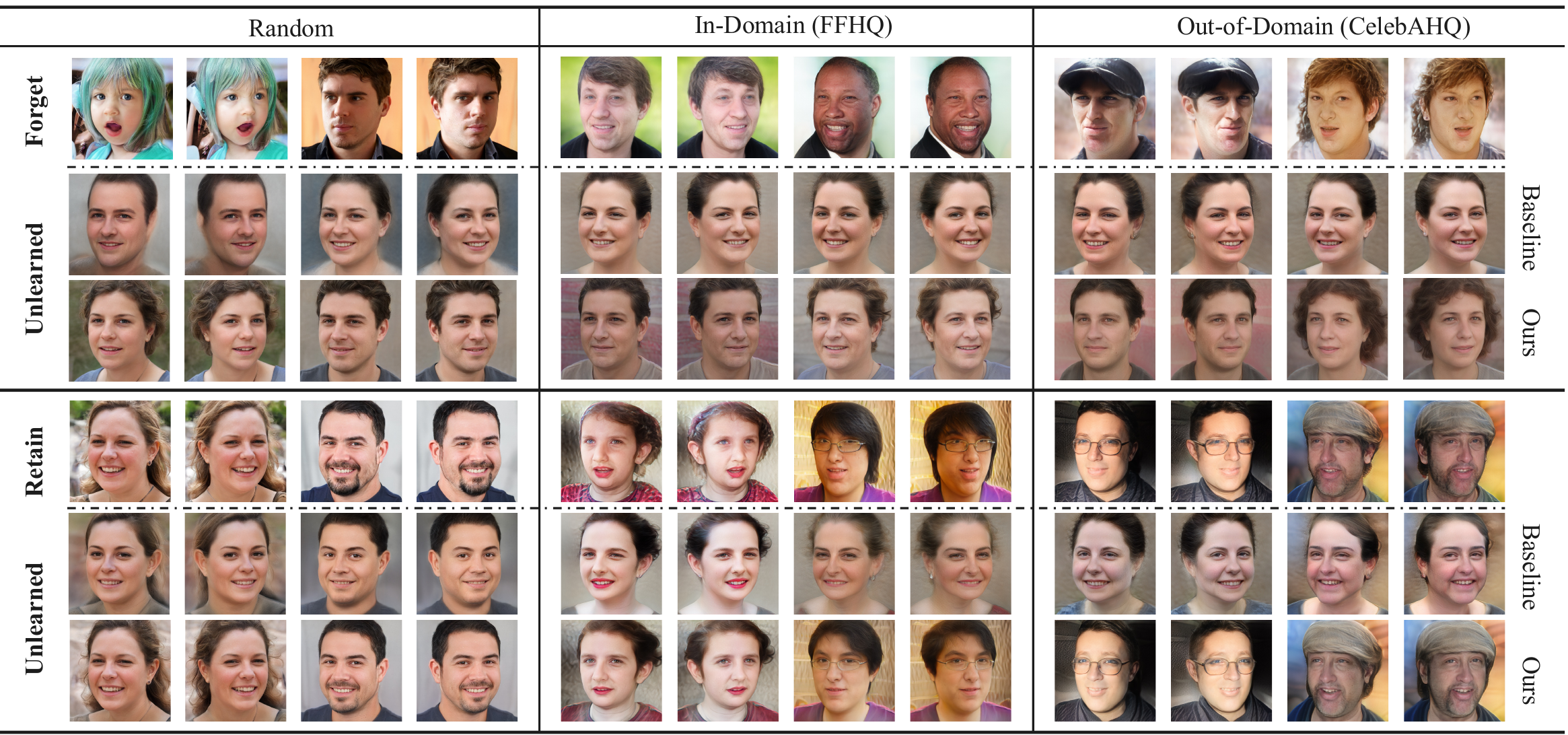}
    \caption{Qualitative results comparing our method with the baseline for the unlearning multiple identities task. The first block illustrates identities targeted for forgetting, while the second block shows identities to be retained. Within each block, the first row contains the original images, while the second and third rows present the generated images corresponding to these identities.}
    \label{fig:quali-forget-retain-main}
    \vspace{-0.2cm}
\end{figure*}
\subsection{Experimental Setup}
\smbf{Datasets.} Following the settings of Seo et al.~\cite{seo2024generative}, we evaluated GUIDE in three forgetting scenarios: Random, InD (in-domain), and OOD (out-of-domain). 
These scenarios represent different sources of identities to unlearn. 
Specifically, identities are sampled randomly from the latent space (Random), from the FFHQ dataset~\cite{karras2019style} used for pre-training (InD), or from the CelebA-HQ dataset~\cite{karras2018progressive} as unlearning targets (OOD). 
For both in-distribution (InD) and out-of-distribution (OOD) settings, we use a GAN inversion network $\mathcal{I}$ to map each forgetting image $x_u^i \in \mathcal{D}_u$ to a latent code $w_u^i=E_{\psi}(x_u^i)$, forming the forget set $W_u=\{w_u^i\}_{i=1}^{N}$; the retaining set $W_r$ is sampled from the same datasets and inverted analogously.

\smbf{Baselines. }
We select GUIDE~\cite{seo2024generative} as our baseline for comparison, as it aligns with our focus on identity-specific unlearning. 
GUIDE effectively removes identity-related features while preserving the overall structure and quality of the generated images. 
In contrast, other methods, such as Selective Amnesia (SA)~\cite{heng2024selective} and Feng et al.~\cite{feng2025controllable}, map forgotten identities to Gaussian noise. 
This approach often results in unnatural artifacts and disrupts facial reconstruction, making it unsuitable for our task. 
Specifically, mapping to Gaussian noise prevents the generation of coherent and realistic reconstructions after unlearning, which is a critical requirement in our setting.

\smbf{Evaluation Metrics.} 
We use Identity Similarity (\idMetric{}), computed with the face recognition network CurricularFace~\cite{huang2020curricularface}, to measure the similarity between images of faces generated from the same latent codes before and after unlearning. 
Additionally, we use the Fr\'{e}chet Inception Distance (\fidMetric{}) score~\cite{heusel2017gans} to quantify the overall visual quality and distribution alignment of the generated images. 
These metrics are calculated on both the forgetting set $\mathcal{D}_u$ and the retaining set $\mathcal{D}_r$, where $\mathcal{D}_r$ is sampled from the same datasets to which $\mathcal{D}_u$ belongs.

\subsection{Experimental Results}
\subsubsection{Balancing Forgetting and Retaining}
\begin{table*}[bth!]
\centering
\caption{Quantitative comparison of model utility in generating \textbf{unforgotten} identities between our method and the baseline (GUIDE) in the generative identity unlearning task, as the number of forgotten identities ($N$) increases.}
\label{tab:main-retain}
\resizebox{0.85\textwidth}{!}{
\begin{tabular}{@{}ll|ll|ll|ll@{}}
\toprule
\multirow{2}{*}{\#IDS} & \multirow{2}{*}{Methods} & \multicolumn{2}{c}{\makecell{FFHQ \\(In-domain Distribution)}}                                    & \multicolumn{2}{c}{\makecell{CelebAHQ \\(OOD Distribution)}}  & \multicolumn{2}{c}{Random}                       \\ \cmidrule(lr){3-4}\cmidrule(lr){5-6}\cmidrule(lr){7-8} 
                       &                          & \multicolumn{1}{c}{\idMetric{} $(\uparrow)$}                                                     & \multicolumn{1}{c}{\fidMetric{} $(\downarrow)$} & \multicolumn{1}{c}{\idMetric{} $(\uparrow)$} & \multicolumn{1}{c}{\fidMetric{} $(\downarrow)$} & \multicolumn{1}{c}{\idMetric{} $(\uparrow)$} & \multicolumn{1}{c}{\fidMetric{} $(\downarrow)$} \\ \cmidrule{1-8}
\multirow{2}{*}{N=1}   & GUIDE                    
& 0.3301 \scriptsize\text{ $\pm$0.0073} 
& 112.06 \scriptsize\text{ $\pm$5.1507}           
& 0.5118 \scriptsize\text{ $\pm$0.0027}          
& 133.16 \scriptsize\text{ $\pm$4.9765}         
& 0.6061 \scriptsize\text{ $\pm$0.0061}          
& 66.531 \scriptsize\text{ $\pm$2.4249}                  \\
  & Ours                     
  & 0.5730 \scriptsize\text{ $\pm$0.0103}   
  & 94.873 \scriptsize\text{ $\pm$0.3855}          
  & 0.7057 \scriptsize\text{ $\pm$0.0055}         
  & 66.280 \scriptsize\text{ $\pm$4.3198}          
  & 0.6457 \scriptsize\text{ $\pm$0.0030}        
  & 87.257 \scriptsize\text{ $\pm$6.9018}          \\ 
  \cmidrule{1-8}

\multirow{2}{*}{N=5}   
& GUIDE                    
& 0.2233 \scriptsize\text{ $\pm$0.0040}                                             
& 143.23 \scriptsize\text{ $\pm$1.4068}          
& 0.3244 \scriptsize\text{ $\pm$0.0292}         
& 116.47 \scriptsize\text{ $\pm$2.0877}         
& 0.4346 \scriptsize\text{ $\pm$0.0019}          
& 144.40 \scriptsize\text{ $\pm$0.1790}         
\\
& Ours                     
& 0.5500 \scriptsize\text{ $\pm$0.0022}                      
& 113.10 \scriptsize\text{ $\pm$0.9406}           
& 0.5766 \scriptsize\text{ $\pm$0.0070}          
& 101.78 \scriptsize\text{ $\pm$2.6185}          
& 0.6538 \scriptsize\text{ $\pm$0.0017}          
& 115.53 \scriptsize\text{ $\pm$0.5600}         \\ 
\cmidrule{1-8}
\multirow{2}{*}{N=10}  
& GUIDE                    
& 0.2132 \scriptsize\text{ $\pm$0.0063}                    
& 181.66 \scriptsize\text{ $\pm$2.5615}           
& 0.2893 \scriptsize\text{ $\pm$0.0024}         
& 165.37 \scriptsize\text{ $\pm$0.8487}         
& 0.2798 \scriptsize\text{ $\pm$0.0012}          
& 138.19 \scriptsize\text{ $\pm$0.6920}         \\
& Ours                     
& 0.5001 \scriptsize\text{ $\pm$0.0030}                     
& 140.86 \scriptsize\text{ $\pm$2.3990}            
& 0.4791 \scriptsize\text{ $\pm$0.0060}          
& 118.32 \scriptsize\text{ $\pm$0.9936}        
& 0.4392 \scriptsize\text{ $\pm$0.0010}          
& 98.023 \scriptsize\text{ $\pm$0.4619}          \\ 
\cmidrule{1-8}
\multirow{2}{*}{N=20}  & GUIDE                    
& 0.1150 \scriptsize\text{ $\pm$0.0036}                     
& 191.50 \scriptsize\text{ $\pm$1.5908}           
& 0.1748 \scriptsize\text{ $\pm$0.0116}         
& 168.34 \scriptsize\text{ $\pm$2.5693}         
& 0.2281  \scriptsize\text{ $\pm$0.0007}          
& 155.08  \scriptsize\text{ $\pm$0.6582}         \\
& Ours                     
& 0.4515 \scriptsize\text{ $\pm$0.0051}                  
& 149.56 \scriptsize\text{ $\pm$1.1429}           
& 0.4182 \scriptsize\text{ $\pm$0.0051}         
& 124.31 \scriptsize\text{ $\pm$5.2634}         
& 0.4774 \scriptsize\text{ $\pm$0.0024}          
& 90.080 \scriptsize\text{ $\pm$0.1015}          \\ 
\cmidrule{1-8}
\multirow{2}{*}{N=50}  
& GUIDE                    
& 0.0115 \scriptsize\text{ $\pm$0.0112}                    
& 197.91 \scriptsize\text{ $\pm$0.5813}           
& 0.1339 \scriptsize\text{ $\pm$0.0054}        
& 174.96 \scriptsize\text{ $\pm$3.2256}         
& 0.1569 \scriptsize\text{ $\pm$0.0005}          
& 141.61 \scriptsize\text{ $\pm$0.3855}         \\
& Ours                     
& 0.3583  \scriptsize\text{ $\pm$0.0043}                    
& 164.07  \scriptsize\text{ $\pm$1.8977}           
& 0.3504  \scriptsize\text{ $\pm$0.0043}        
& 137.41  \scriptsize\text{ $\pm$1.9314}         
& 0.4612  \scriptsize\text{ $\pm$0.0009}        
& 88.306  \scriptsize\text{ $\pm$0.0092} \\
\midrule
\multicolumn{2}{c|}{$ \textit{Average Impr. (\%)}$} 
& \posNum{96.424}		
& \posNum{19.568}						
& \posNum{732.18}	
& \posNum{27.780}
& \posNum{83.586}						
& \posNum{18.928} \\ 
\bottomrule
\end{tabular}
}
\vspace{-0.5cm}
\end{table*}

To evaluate the ability of~\methodName{} to balance the trade-off between forgetting and retaining in unlearning tasks, we conducted experiments with an increasing number of forgotten identities, where $K \in \{1, 5, 10, 20, 50\}$, across three categories: \textit{In-domain Distribution}, \textit{OOD Distribution}, and \textit{Random}. We then compare our method with the baseline both quantitatively and qualitatively, and incorporate human judgment.

\smbf{Enhancing Retained Model Utility. }
First, we report the improvement of our method in maintaining model utility in~\autoref{tab:main-retain} in the generative identity unlearning task across three categories.
Our method consistently outperforms the baseline (GUIDE) in maintaining higher \idMetric{} scores and lower \fidMetric{} values for retained identities, even in scenarios where only one identity is forgotten. 
As the number of forgotten identities ($N$) increases, unlearning becomes more challenging, leading to lower identity similarity and higher image distortion. 
However, our method still achieves better performance, with \idMetric{} scores improving by 50--150\% over GUIDE, especially in the CelebAHQ dataset, where \idMetric{} scores are roughly 2x higher for $N=1$ to $N=20$. 
Additionally, our method demonstrates a 
substantial
reduction in \fidMetric{} values, highlighting its ability to preserve image quality while unlearning more identities.
\textbf{On average, our method can improve the retainability of the model by up to 700\% across these scenarios. }
This observation is represented clearly in the qualitative results in~\autoref{fig:quali-forget-retain-main}. 
The images in the second row exhibit visually apparent distortions, suggesting that the baseline approach inadvertently removes crucial facial attributes beyond the intended identity modification. 
In contrast, our method, shown in the third row, maintains a higher level of visual consistency, preserving defining characteristics such as facial structure, expression, and texture. 
This indicates that the proposed approach is more effective in unlearning forgotten identities without dramatically affecting the retained identities. 

\smbf{Ensuring Identity Forgetting. }
To evaluate the model’s performance in effectively unlearning identities, we present quantitative results in~\autoref{tab:main-forget}. On average, \idMetric{} scores for forgotten identities are 54.7\% lower than those for retained identities, indicating that \methodName{} effectively prevents the generation of images from forgotten identities. We argue that forgetting an identity does not require aggressively minimizing metrics like \idMetric{}, as this could harm utility, resulting in the degraded performance demonstrated in~\autoref{fig:quali-forget-retain-main}.
This raises the question of when an identity can be considered sufficiently forgotten. Since there is no clear threshold for \idMetric{} scores at which humans perceive an unlearned image as distinct rather than identical, determining sufficiency is inherently subjective. To address this, we conducted a human recognition study to assess (i) whether unlearned identities are no longer identifiable by humans and (ii) whether retained identities remain recognizable. This evaluation provides a more robust, quantitatively grounded assessment of each model's effectiveness in identity unlearning beyond relying solely on \idMetric{} scores.
\begin{table}[t!]
\centering
\caption{Quantitative comparison of the forgetting ability of the unlearned model, measured on \textbf{forgotten} identities, between our method and the baseline (GUIDE) in the generative identity unlearning task, as the number of forgotten identities ($N$) increases.}
\label{tab:main-forget}
\resizebox{0.90\linewidth}{!}{
\begin{tabular}{@{}clccccc@{}}
\toprule
\multirow{2}{*}{Metric} & \multirow{2}{*}{Model} & \multicolumn{5}{c}{Number of Forgetting IDs}  \\ \cmidrule(lr){3-7} 
                        &                        
                        & N=1                
                        & N=5 
                        & N=10 
                        & N=20 
                        & N=50 \\ \cmidrule(lr){1-7}
\multirow{2}{*}{\idMetric{}}     & GUIDE                  
& \makecell{0.2773 \\ \small\textit{ \quad\quad$\pm$0.0099}} 
&  \makecell{0.0040 \\ \small\textit{ \quad\quad$\pm$0.0077} }    
&  \makecell{0.0095 \\ \small\textit{ \quad\quad$\pm$0.0017}  }    
&  \makecell{-0.0275 \\ \small\textit{ \quad\quad$\pm$0.0034}  }    
&  \makecell{-0.0431 \\ \small\textit{ \quad\quad$\pm$0.0018}   }    \\
& Ours                   
&  \makecell{0.3576 \\\small\textit{ \quad\quad$\pm$0.0146}      }              
&  \makecell{0.2664 \\ \small\textit{ \quad\quad$\pm$0.0134}}     
&  \makecell{0.2559 \\ \small\textit{ \quad\quad$\pm$0.0078}}      
&  \makecell{0.2917  \\ \small\textit{ \quad\quad$\pm$0.0050}}      
&  \makecell{0.2433 \\ \small\textit{ \quad\quad$\pm$0.0051}}      \\ \cmidrule(lr){1-7}
\multirow{2}{*}{\fidMetric{}}    
& GUIDE                  
&  \makecell{133.90 \\ \small\textit{ \quad\quad$\pm$6.9372}}                  
&  \makecell{177.27 \\ \small\textit{ \quad\quad$\pm$2.2242}}     
&  \makecell{163.83 \\ \small\textit{ \quad\quad$\pm$0.9667}}      
&  \makecell{143.44 \\ \small\textit{ \quad\quad$\pm$0.8456}}      
&  \makecell{146.43 \\ \small\textit{ \quad\quad$\pm$0.6706}}      \\
& Ours 
&  \makecell{115.98 \\ \small\textit{ \quad\quad$\pm$7.7846}}
&  \makecell{125.98 \\ \small\textit{ \quad\quad$\pm$0.8796}}                  
&  \makecell{125.93 \\ \small\textit{ \quad\quad$\pm$0.3963}}     
&  \makecell{105.40 \\ \small\textit{ \quad\quad$\pm$0.6074}}         
&  \makecell{106.12 \\ \small\textit{ \quad\quad$\pm$1.4283}}      \\ 
                        \bottomrule
\end{tabular}
}
\vspace{-0.4cm}
\end{table}

\begin{figure}[tb!]
    \centering
    \begin{subfigure}{0.62\linewidth}
        \includegraphics[width=0.95\linewidth]{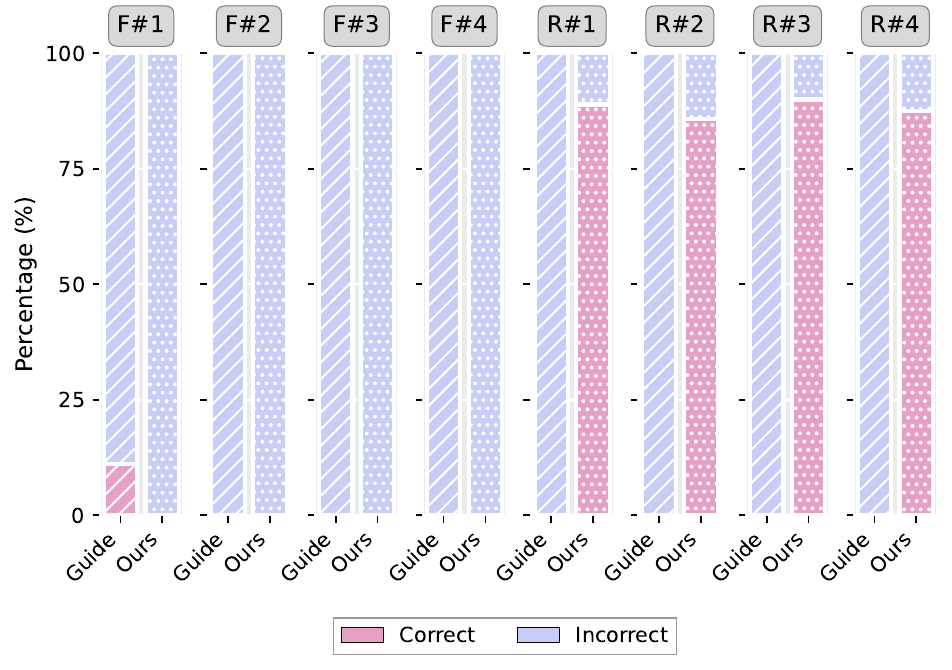}
        \caption{Accuracy of Selected Identities}
        \label{fig:human-study-individual}
    \end{subfigure}
    \begin{subfigure}{0.34\linewidth}
        \includegraphics[width=0.95\linewidth]{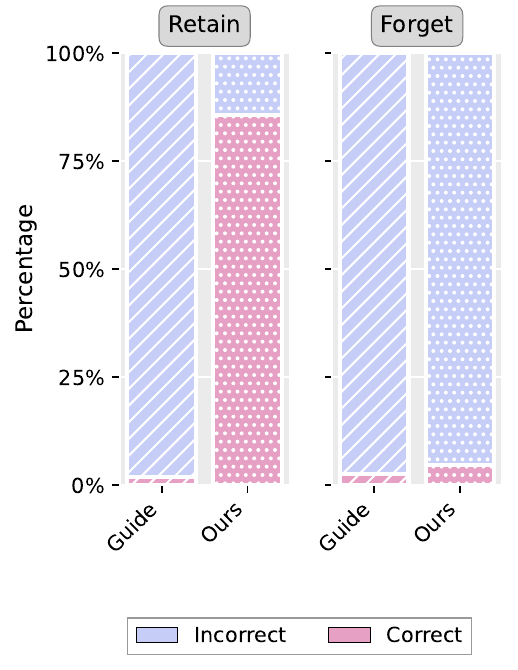}
        \caption{Overall Accuracy}
    \end{subfigure}
    \caption{Human judgment result on selected forgetting (\texttt{F\#}) and retaining (\texttt{R\#)} identities.}
    \label{fig:human-study}
    \vspace{-0.5cm}
\end{figure}

\smbf{Human Judgments' Results. }
In this study, we sample the generated images for 20 forgotten identities and 20 retaining identities produced by our unlearned model and GUIDE's unlearned model. 
We recruited 60 participants and received 579 responses. Each participant was asked to select the image among five sampled images that they believe depicts the same person given a set of five random identities' images.
Interestingly, participants were unable to consistently identify the original identity among the unlearned images for either GUIDE or \methodName{}, demonstrating comparable effectiveness in identity forgetting. As shown in~\autoref{fig:human-study}, performance on the forgetting set is indistinguishable between the two methods. The individual breakdown in~\autoref{fig:human-study-individual} reveals that 80–100\% of participants failed to match original images to any unlearned outputs.  
However, for retained identities, 
Specifically, participants correctly identified \methodName{}-generated identities 86\% of the time, whereas for GUIDE, this rate is only 1.9\%. These results highlight that \textbf{\methodName{} effectively balances identity forgetting while preserving overall model utility.}

\section{Ablation Study}
In this section, we conduct ablation studies to evaluate controllable unlearning, identity retention, global utility preservation, and sequential unlearning.
\begin{figure}[tbh!]
    \centering
    \includegraphics[width=0.85\linewidth]{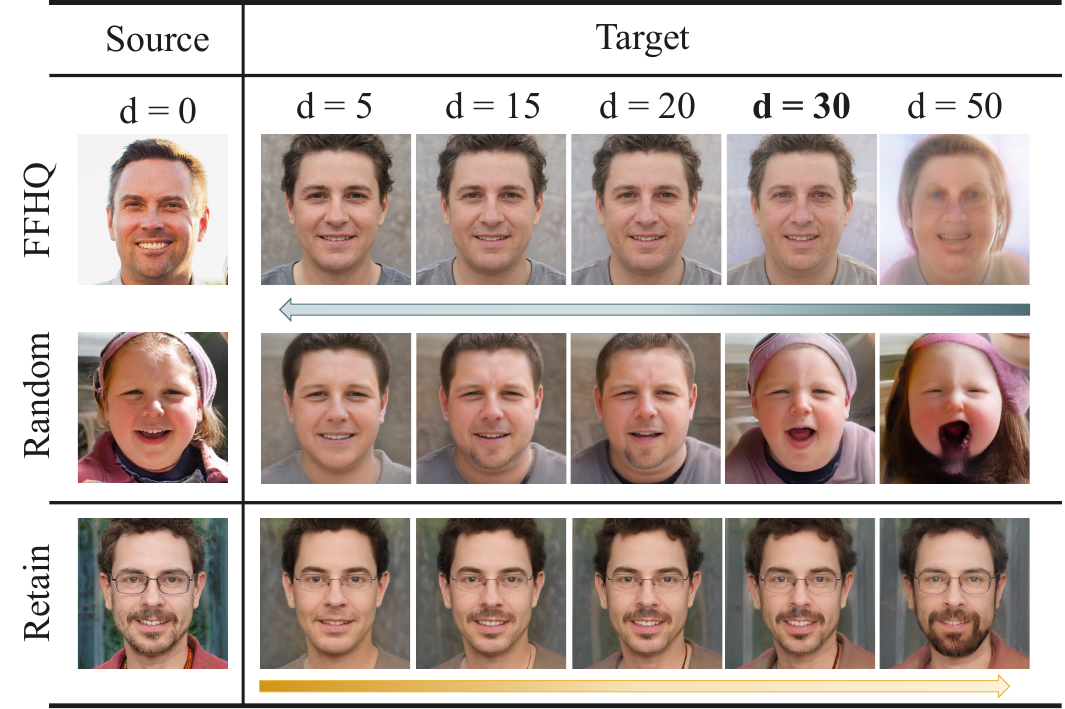}
    \caption{Visualization of our controllable unlearning.  The effect of unlearning in I2I generative models is controlled by $d$ (\autoref{eqn:new-id}) for the vectors generated by our de-identification models. With our approach, the model vendor can selectively control the forgetting level. The higher the value of $d$ (i.e., up to 30), the more original features of forgetting people are retained.}
    \label{fig:change-d-target}
    \vspace{-0.3cm}
\end{figure}

\noindent\textbf{$d-$Controllable Unlearning.} Our method can be used as a controllable unlearning solution in I2I generative models, where $d$ is the control coefficient. In \autoref{eqn:transformation}, with a well-trained model $\Phi$, the targeted identity for each identity $w_{id}$ is adjusted with a different coefficient $d$. As shown in \autoref{fig:change-d-target}, the higher the value of $d$, the more features of the forgetting ID $w_{id}$ are retained in the new mapping identity $w_t$. 
With our method, the model vendor can select this controllable parameter to balance the trade-off and the forgetting level that reaches a consensus between the vendor and the requested unlearning party. 
However, there is a trade-off between forgetting and retaining utility, as discussed in previous work~\cite{feng2025controllable, zhang2023review}. This relationship is presented in \autoref{tab:change-target-d}. While reducing the distance can make the model forget the forgetting set $\mathcal{D}_f$ more drastically (i.e., lower \idMetric{} and higher \fidMetric{} for the forgetting set), this leads to a greater impact on the retaining set (i.e., lower \idMetric{}). 
It is worth noting that even with $d=30$, the generated images are much different from the original identities.
\begin{table}[tb!]
\centering
\caption{Ablation study on the effect of $d$-controllable value in balancing forgetting and retaining capabilities. We used FFHQ and Random datasets in this experiment.}
\label{tab:change-target-d}
\resizebox{0.95\linewidth}{!}{
\begin{tabular}{@{}clccccccc@{}}
\toprule
\multirow{2}{*}{Metric} & \multirow{2}{*}{Dataset} & \multicolumn{7}{c}{Value of $d$}                             \\ \cmidrule(lr){3-9} 
                        &                          & d=-5   & d=5    & d=10   & d=15   & d=20   & d=30   & d=50   \\ \cmidrule(lr){1-9}
\multirow{2}{*}{\idMetric{}}     & Forget                   & 0.1815 & 0.2408 & 0.2713 & 0.3194 & 0.3904 & 0.4472 & 0.5124 \\
                        & Retain                   & 0.4501 & 0.4960 & 0.5107 & 0.5542 & 0.5880 & 0.7073 & 0.5949 \\ \bottomrule
\end{tabular}}
\vspace{-0.5cm}
\end{table}

\begin{figure*}[tbh!]
    \centering
    \begin{minipage}{.27\linewidth}
        \includegraphics[width=0.9\linewidth]{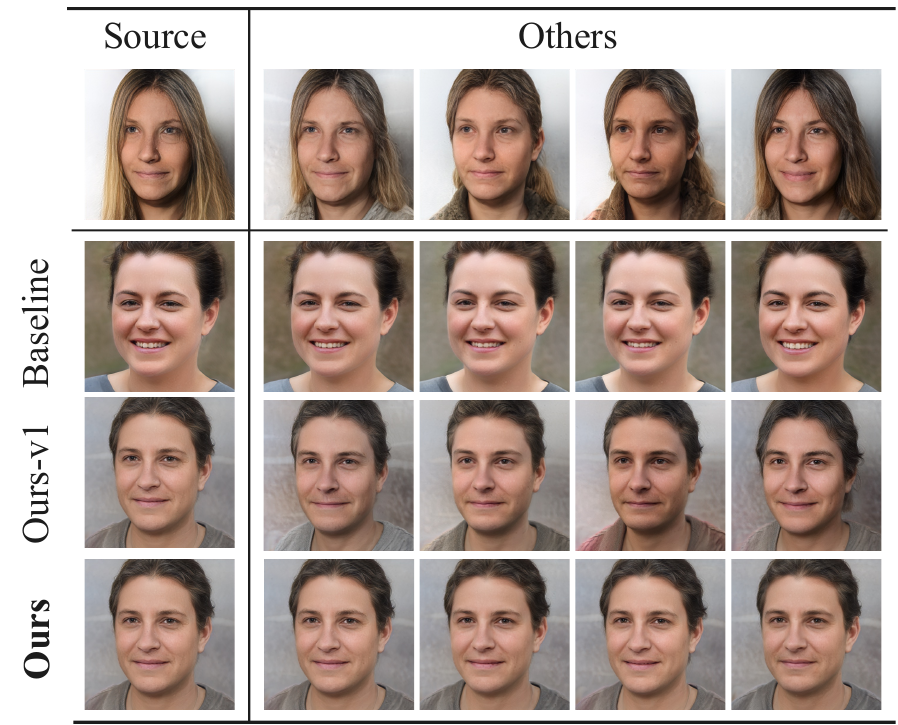}
        \caption{Reconstructed images for the forgotten identity and their unseen neighbor samples (Others) from the same person to demonstrate the thoroughness of identity removal. 
``Ours-v1'' denotes the variant without the neighbor (vicinity) loss from \autoref{eqn:forget-loss}.}
    \label{fig:remove-l-adj}
    \vspace{-0.3cm}
    \end{minipage}\hfill
    \begin{minipage}{.41\linewidth}
        \includegraphics[width=1.0\linewidth]{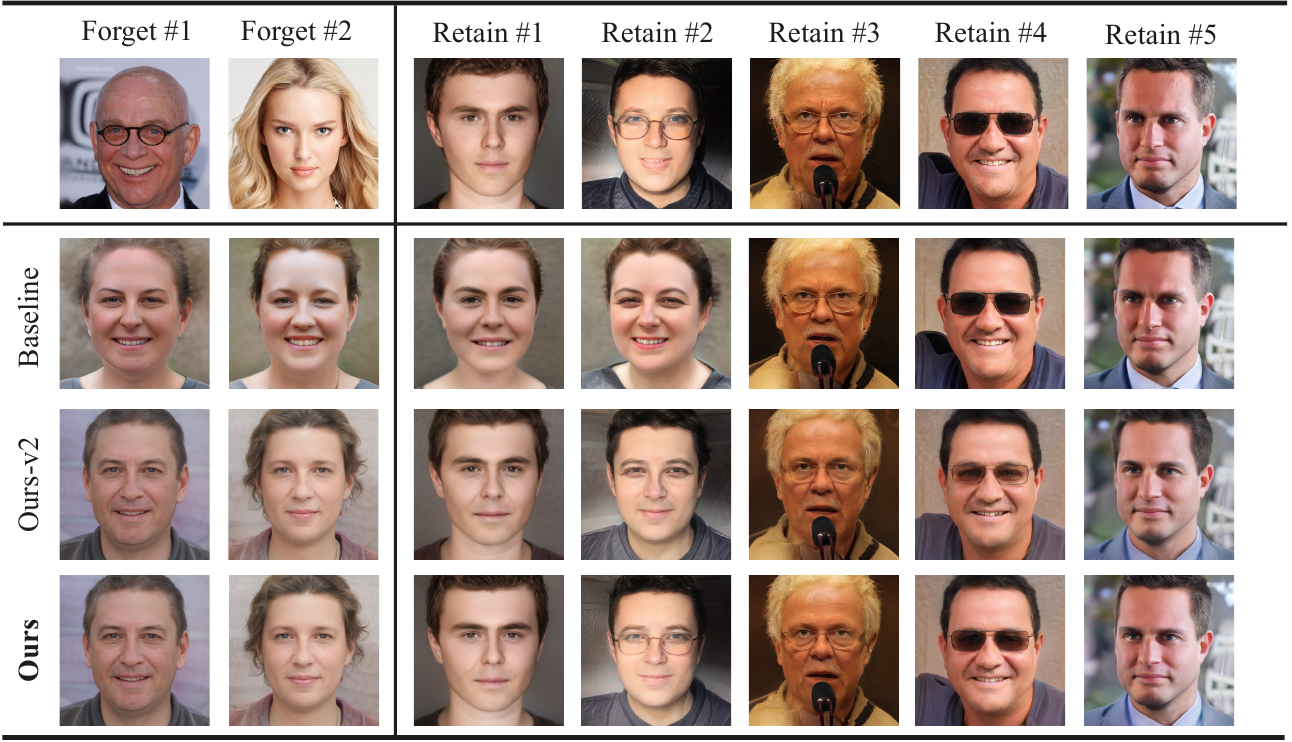}
        \caption{Qualitative results when removing the $\mathcal{L}_{ewc}$ term from the full objective (Eq.~\ref{eqn:loss-full}); we denote this variant as \emph{Ours-v2}. The full model (with $\mathcal{L}_{ewc}$) better preserves retained identities, maintaining fine-grained details (e.g., glasses, hair, wrinkles).}
    \label{fig:remove-l-glob}
    \vspace{-0.3cm}
    \end{minipage}\hfill
    \begin{minipage}{.3\linewidth}
        \includegraphics[width=0.9\linewidth]{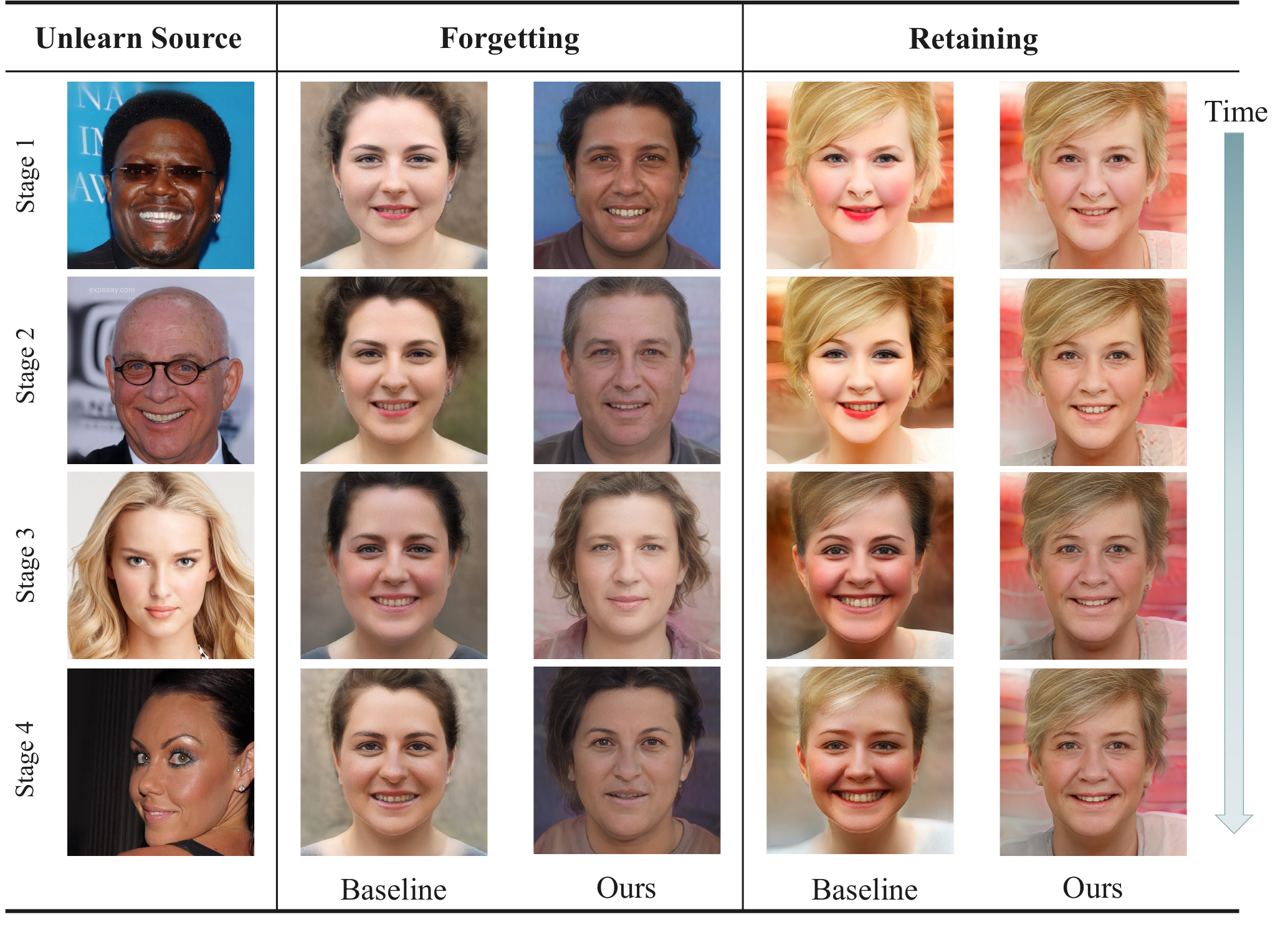}
            \caption{Qualitative results for sequential unlearning setting from stage 1 to stage 4: in each stage, we compare the reconstructed images for both forgetting and retaining identities. \textit{The resulting model of previous unlearning stage is used for the next unlearning stage.}}
    \label{fig:sequential-unlearn}
    \vspace{-0.3cm}
    \end{minipage}
\end{figure*}

\noindent\textbf{Dual Objective: Identity Removal and Global Utility Preservation. }
We study the impact of our two loss terms: forgetting neighbor loss $\mathcal{L}_{nei}$ for thorough identity forgetting and the global preservation loss $\mathcal{L}_{\text{ewc}}$ for utility retention. To evaluate identity removal, we conduct a multi-image test on CelebAHQ identities, assessing whether our method forgets other images of the same identity that are not explicitly included in $W_u$.
\autoref{fig:remove-l-adj} summarizes the results. ``Ours-v1'' denotes the variant \emph{without} the neighbor loss $\mathcal{L}_{nei}$ in \autoref{eqn:forget-loss}, so the forgetting objective applies only to the anchor images of each target identity. Without this term, residual attributes of the forgotten identity can persist, even when the generated images depict distinct individuals. Adding $\mathcal{L}_{nei}$ enforces consistency across the local neighborhood in latent space, which is crucial for the ultimate goal of removing an entire identity rather than only a single image.
To analyze $\mathcal{L}_{ewc}$, \autoref{fig:remove-l-glob} compares reconstructions with (Ours) and without (Ours-v2) the global preservation loss. The first two retained identities come from the same distribution as the forgotten ones, while the last three (IDs 3–5) are randomly sampled latents. Random identities remain largely unaffected by unlearning (even under GUIDE), whereas closer identities (IDs 1–2) are more susceptible to collateral forgetting. Ours-v2 already preserves these better than GUIDE due to our de-identification, and adding $\mathcal{L}_{ewc}$ further improves retention by preserving fine details (e.g., glasses, hair, wrinkles).
We provide a detailed quantitative result for our ablation study in Appendices B.1. and B.2.

\noindent\textbf{Sequential unlearning.}
We consider the setting where unlearning requests arrive \emph{sequentially} rather than in a single batch. Let $G_{\theta}^{(0)} \!=\! G_{\theta}^*$ be the pretrained model and let $\mathcal{D}_f^{(k)}$ denote the (new) identities to forget at stage $k{=}1,\dots,4$. At each stage, we initialize from the previously unlearned model and run the \emph{same} unlearning procedure (architecture, losses, schedule, and hyperparameters unchanged):
\(G_{\theta}^{(k)} \;\leftarrow\; \textsc{Unlearn}\!\big(G_{\theta}^{(k-1)},\, \mathcal{D}_f^{(k)}\big).
\)
Figure~\ref{fig:sequential-unlearn} shows qualitative results after four stages for GUIDE and \methodName{}. While GUIDE accumulates collateral damage—progressively eroding features of the \emph{retained} identities across stages—\methodName{} continues to forget the requested identities while preserving the appearance of the retained set. Detailed setup and additional analysis are provided in Appendix B.5.

\paragraph{Privacy Discussion}
\begin{figure}[bth!]
    \vspace{-0.5cm}
    \begin{minipage}{0.4\linewidth} %
        \centering
        \includegraphics[width=0.99\linewidth]{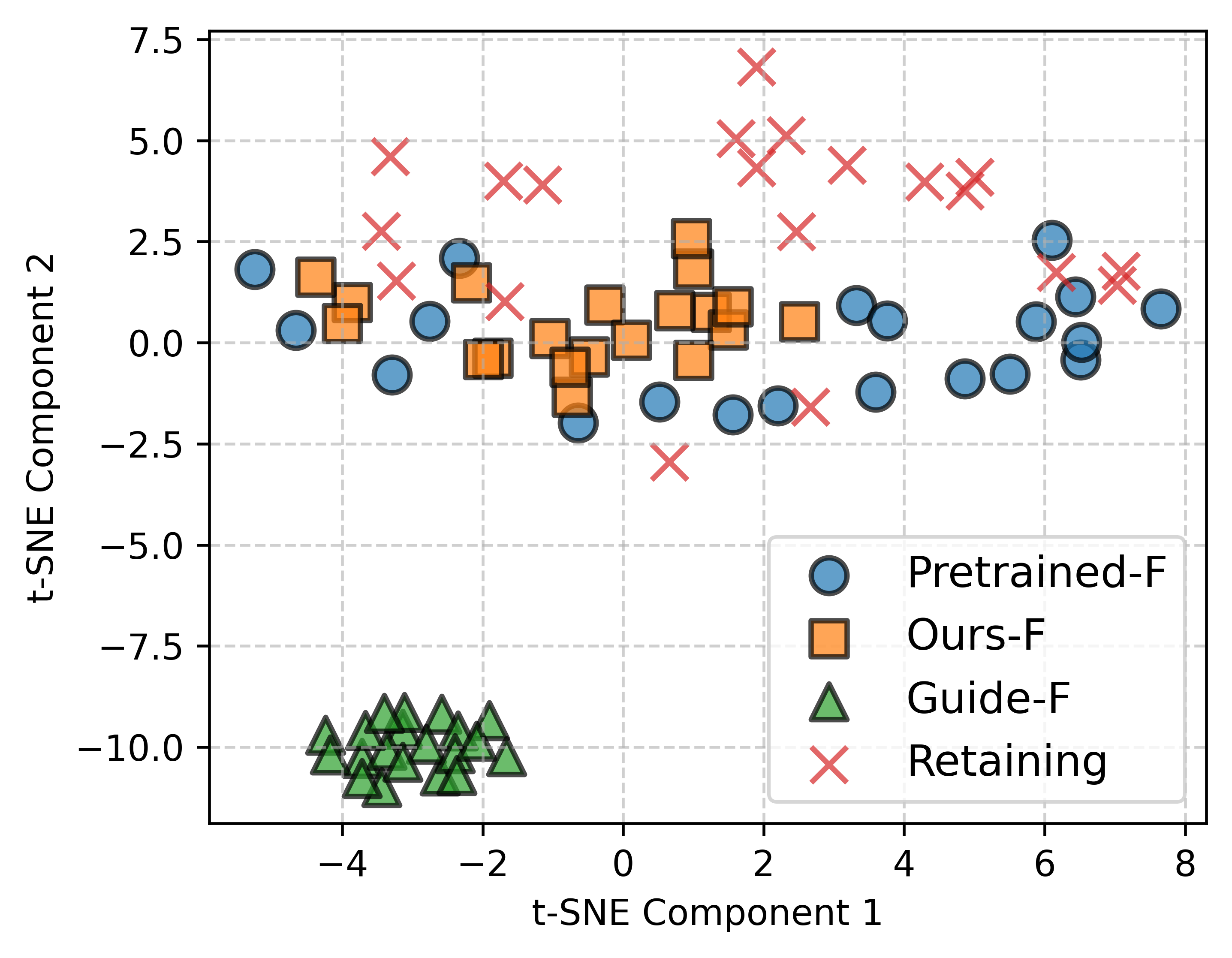}
    \end{minipage}
    \hfill
    \begin{minipage}{0.55\linewidth}
    \captionof{table}{Quantitative Results}
\label{tab:security}
\vspace{-0.2cm}
\resizebox{0.95\linewidth}{!}{
\begin{tabular}{@{}llcccc@{}}
\toprule
\multirow{2}{*}{}               & \multirow{2}{*}{Methods} & \multicolumn{4}{c}{Metrics} \\ \cmidrule(l){3-6} 
&                          
& MSE    
& PSNR   
& LPIPS   & SSIM   \\ \cmidrule(r){1-6}
\multirow{2}{*}{\rotatebox{90}{Forgetting}} 
& GUIDE                    
& \makecell{254.31\\ \scriptsize\textit{$\pm$ 376.36}}        
& \makecell{25.4262\\ \scriptsize\textit{$\pm$ 0.2026}}        
& \makecell{0.2048\\ \scriptsize\textit{$\pm$ 0.0000}}         
& \makecell{25.4262\\ \scriptsize\textit{$\pm$ 0.2026}}       \\ \cmidrule(r){2-6}
& Ours                     
& \makecell{1022.81\\ \scriptsize\textit{55301.1}}     
& \makecell{18.5459\\ \scriptsize\textit{$\pm$ 0.8418}}         
& \makecell{0.2936\\ \scriptsize\textit{$\pm$ 0.0002}}         
& \makecell{0.7156\\ \scriptsize\textit{$\pm$ 0.0002}}       \\ 
                                \bottomrule
\end{tabular}}

    \end{minipage}
    \caption{Privacy analysis of images
generated by GUIDE and~\methodName{}. The left figure is a t-SNE visualization of \update{intermediate tri-plane representation} from forgetting (\texttt{-F}) and retaining identities; from the pre-trained model (before unlearning), the unlearned model produced by~\methodName{}, and the baseline GUIDE. 
    }
    \label{fig:tsne-forget-retain}
    \vspace{-0.3cm}
\end{figure}%
In this section, we assess the security of \methodName{} in concealing forgotten identities. 
Prior unlearning methods (e.g., GUIDE, Selective Amnesia) collapse erased identities to an average face or noise, creating a recognizable \emph{signature} that enables simple erasure-detection and membership-inference attacks (e.g., nearest-centroid tests, one-class density checks) by pushing outputs into low-density, out-of-manifold regions. In contrast, \methodName{} maps each erased identity to a \emph{close, diverse counterfactual} that remains on the data manifold, eliminating a single surrogate template and reducing separability from natural variation. Empirically, Table~\ref{tab:security} shows larger divergence from each identity’s pretrained counterpart yet higher \emph{intra-forgotten} variability (higher MSE/LPIPS, lower PSNR/SSIM) than GUIDE, which makes unsupervised identification harder; \autoref{fig:tsne-forget-retain} likewise shows GUIDE’s forgotten samples forming a distinct cluster, while \methodName{} intermixes with retained samples. This distributional indistinguishability limits adaptive ``probe-and-collapse'' and template-matching attacks, preserves fidelity for retained identities (avoiding collateral degradation seen with noise/mean collapse), and scales securely since a single counterfactual generator yields varied surrogates without a deterministic average-face cue. To this end, these properties make \methodName{} more secure and privacy-preserving than average/noise-based unlearning, challenging membership inference attacks.

\section{Conclusion}

In this work, we introduce~\methodName{}, an approach for effectively unlearning multiple identities from a generative model, either simultaneously or sequentially.
Unlike existing methods, \methodName{} autonomously determines optimal target representations for each forgotten identity, ensuring a structured and efficient continual unlearning process.
Through qualitative and quantitative evaluations, we demonstrate that~\methodName{}
outperforms baselines in preserving model utility while achieving effective identity removal.
Furthermore, our analysis provides deeper insights into how unlearning impacts different regions of the data distribution, offering a foundation for future research on adaptive target determination in unlearning generative models.
\section*{Acknowledgements}
We acknowledge partial support from the National Security Agency under the Science of Security program, from the Defense Advanced Research Projects Agency under the CASTLE program, the Advanced Research Projects Agency for Health, and from an Amazon Research Award. The contents of this paper do not necessarily reflect the views of the US Government.
{
    \small
    \bibliographystyle{ieeenat_fullname}
    \bibliography{main}
}

\end{document}